# What is usual in unusual videos?
# Trajectory snippet histograms for discovering unusualness

Ahmet Iscen
Bilkent University
Ankara, Turkey 06800
ahmet.iscen@bilkent.edu.tr

Anil Armagan
Bilkent University
Ankara, Turkey 06800
anil.armagan@bilkent.edu.tr

Pinar Duygulu
Bilkent University
Ankara, Turkey 06800
duygulu@cs.bilkent.edu.tr

## Abstract

*Unusual events are important as being possible indicators of undesired consequences. Moreover, unusualness in everyday life activities may also be amusing to watch as proven by the popularity of such videos shared in social media. Discovery of unusual events in videos is generally attacked as a problem of finding usual patterns, and then separating the ones that do not resemble them. In this study, we address the problem from a different perspective, and try to answer what type of patterns are shared among unusual videos that make them resemble to each other regardless of the ongoing event. With this challenging problem at hand, we propose a novel descriptor to encode the rapid motions in videos utilizing densely extracted trajectories. The proposed descriptor, which is referred to as trajectory snippet histograms, is used to distinguish unusual videos from usual videos, and further exploited to discover snapshots in which unusualness happen. Experiments on domain specific people falling videos and unrestricted funny videos show the effectiveness of our method in capturing unusualness.*

## 1. Introduction

People tend to pay more attention to unusual things and events, and it seems that it is generally amusing to watch them happening as proven by the popularity of TV shows like America's Funniest Home Videos, where video clips with unexpected events are shown. The so called "fail compilations" that refer to the videos that have collections of unusual and funny events are also among the most popular videos shared in social media, such as Youtube or Vine. In spite of their growing amount, there has not been sufficient attention to such videos in computer vision community.

Consider the video frames shown in Figure 1. If a user was presented with these videos, they would probably want to watch the ones on the top row before the ones at the bottom. Yet, what makes these videos more appealing to the audience? The *unusual events* taking place in these videos are likely to have an effect on the preference, compared to the events that we expect to see every day. On the other hand, what makes something unusual? In most of the cases it is difficult to answer this question.

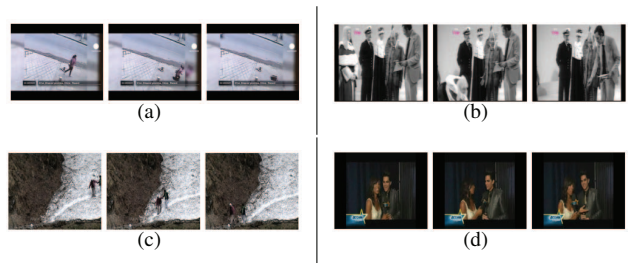

Figure 1. Videos on the top row contain unusual events while the videos on the bottom row do not contain any unusualness. On (a), the subject disappears and falls into the ground while walking, meanwhile the couple on (c) performs a usual walking action without any unexpected events. Similarly, subject standing on (b) collapses during an interview while two subjects on (d) perform a normal interview. Regardless of the action that the subjects are performing, our aim is to distinguish these videos.

In this work, we aim to discover what unusual videos share in common. Our main intuition is that there should be a characteristic motion pattern in such videos, regardless of the ongoing actions and where the event happens. We propose a novel descriptor, which we call *trajectory snippet histograms*, based on the trajectory information of little snippets in each video, and show that it is capable of revealing the differences between *unusual* and *usual* videos. We also use the proposed descriptor to find the discriminative spatio-temporal patches, which we refer to as *snapshots*, that explain what makes these videos unusual.

Although recently the problem of detecting unexpected events has been attacked, the focus is mostly on surveillance videos for capturing specific events in limited domains. Our focus is not to detect the unusual activity in a single video, but rather to capture the common characteristics of being unusual. Moreover, we do not limit ourselves to surveillance videos but rather to the realistic videos shared in social media, in their most natural form with variety of challenges.

The data collected from web is weakly-labeled. While a video in the training set is labeled as *usual* or *unusual*, we do not know which part contains unusualness. We cannot even guarantee that a video labeled as unusual definitely contains an unusual part or a video labeled as usual does not contain an unusual part, since we query based on subjective and noisy user tagging. Our goal in such a setting is to discover the hidden properties of unusual videos from the data itself.

## 2. Related Work

While activity recognition has been a widely studied topic[13], the literature is dominated with the studies on ordinary actions. Some of the early studies that attack the the problem of detecting irregular or unusual activities assume that there are only a few regular activities [23, 1]. However, there are various number of activities in real life.

Surveillance videos has been considered in several studies with the aim of preventing undesired events that are usually the unexpected ones. In [15] dominant and anomalous behaviors are detected by utilising a hierarchical codebook of dense spatio-temporal video volumes. In [22] detecting unusual events in video is formulated as a sparse coding problem with an atomically learned event dictionary forming the sparse coding bases. In [17], normal crowd behavior is modelled based on mixtures of dynamic textures, and anomaly is detected as outliers. Recently, prediction based methods gained attention, as in [24] which focuses on predicting people's future locations to avoid robot collusion and [9] which considers effect of physical environment on human actions for activity forecasting. However, most of these methods are limited with domain specific events for surveillance purposes in constrained environments. We are interested in revealing the unusualness in a much more broader domain focusing on web videos that are considered in the literature for complex event detection[18, 14], but not sufficiently for anomaly detection.

Trajectory based features have been shown to be successful in different applications. Recently, in [21] relying on large collections of videos, a simple model of the distribution of expected motions is built using trajectories of keypoints for event prediction. The dense trajectories has been presented in [19] for recognition of complex activities. We extend the use of dense trajectories to detection of unusualness through a novel descriptor that encodes the motion of trajectories.

For finding common and discriminative parts, Singh *et al.* [16] show that one can successfully detect discriminative patches on images with different categories. In [5], Doersch *et al.* extend this idea by using geo-spatial discriminative patches to differentiate images from one city to another. More recently, Jain *et al.* showed in [8] that it is also possible obtain discriminative patches from videos using examplar-SVMs originally proposed in [12]. In [20], a method for temporal commonality discovery is proposed to find the subsequences with similar visual content.

## 3. Method

When huge number of unrestricted web videos are considered, it is difficult, if not impossible, to learn all possible usual events that could happen, and to distinguish unusual events as the ones that are not encountered previously. We attack the problem from a different perspective, and aim to discover the shared characteristics among unusual videos.

Our main intuition is that unusual events contain irregular and fast movements. These are usually resulted from causes such as being scared or surprised, or sudden actions like falling. To capture such rapid motions we exploit dense trajectories as in [19], and propose a new descriptor that encodes the change in the trajectories in short intervals, that we call as *snippets*. In the following, first we summarize how we utilize dense trajectories, and then present our proposed descriptor *trajectory snippet histograms*, followed by description of our method for *snapshot* discovery.

### 3.1. Finding Trajectories

We utilize the method described in [19] to find trajectories. This method samples feature points densely in different spatial scales with a step size of M pixels, where M=8 in our experiments. Sampled points in regions without any structure are removed since it is impossible to track them. Once the dense points are found, optical flow of the video is computed by applying the Farnebäck's method [6]. Median filtering is applied to optical flow field to maintain sharp motion boundaries. Trajectories are tracked upto $D$ frames apart, to limit drift from the original locations. Static trajectories with no motion information or erroneous trajectories with sudden large displacements are removed. Finally, a trajectory with duration $D$ frames is represented as a sequence $T = (P_t, ..., P_{t+D-1})$ where $P_t = (x_t, y_t)$ is the point tracked at frame $t$. Unlike [19] where $D = 15$ to track trajectories for 15 frames, in order to consider trajectories with fast motion, we set $D$ to 5. This length provides a good trade-off between capturing fast motion, and providing sufficiently long trajectories with useful information[7].

### 3.2. Calculating Snippet Histograms

We use the extracted trajectories to encode the motion in short time intervals, namely in *snippets*. Figure 2 depicts the overview of our method. First, for each trajectory $T$, we make use of the length of the trajectory ($l$), variance along x-axis ($v_x$), and variance along y-axis ($v_y$) to encode the motion information for a single trajectory. Trajectories with longer lengths correspond to faster motions, and therefore velocity is encoded with the length of the trajectory in one temporal unit. We combine it with the variance of trajectory along x and y-coordinates, to encode the spatial extension of the motion.

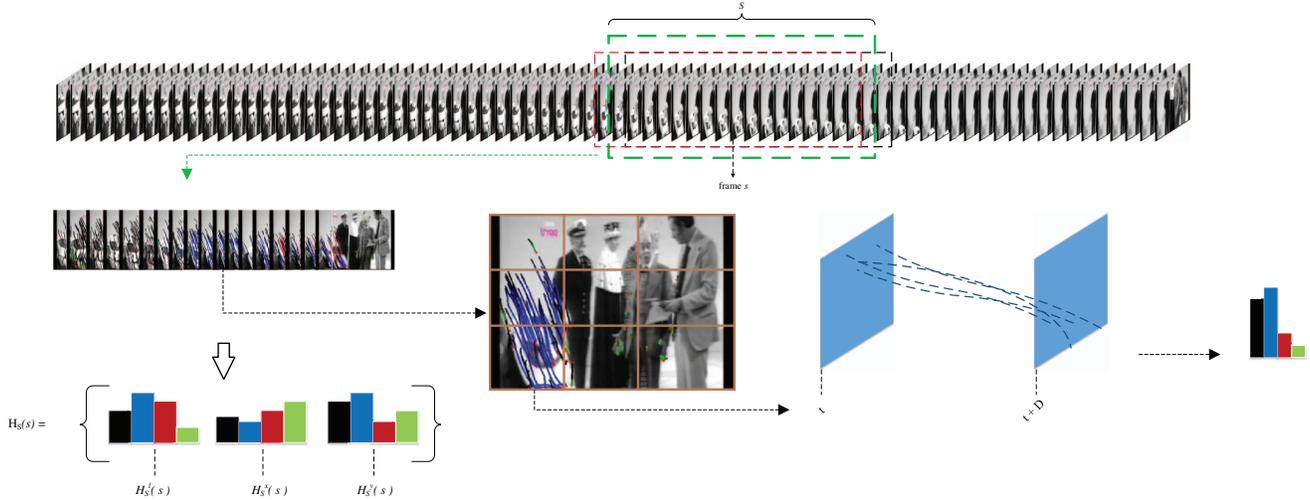

Figure 2. For each snippet $S$ centered at frame $s$ in the video, we extract the trajectory length, variance on x and variance on y values of the frames to construct a histogram of trajectories in snippets. Each frame is divided into $N \times N$ grids, and only trajectories that are centered at those grids contribute to their histogram. This process is repeated for each $s$ in the video in a sliding window fashion.

Let $T$ be a trajectory in a video that starts on frame $t$ and is tracked for a duration of $D$ frames. Let $m_x$ and $m_y$ be the average positions of $T$ on x and y coordinates, respectively. For each trajectory, the variance on x and y coordinates and the length of each trajectory is calculated as in Eq.1.

$$m_x = \frac{1}{D}\sum_t^{t+D-1} x_t, v_x = \frac{1}{D}\sum_t^{t+D-1}(x_t - m_x)^2$$

$$m_y = \frac{1}{D}\sum_t^{t+D-1} y_t, v_y = \frac{1}{D}\sum_t^{t+D-1}(y_t - m_y)^2, \quad (1)$$

$$l = \sum_t^{t+D-1}\sqrt{(x_{t+1} - x_t)^2 + (y_{t+1} - y_t)^2}$$

Note that, videos that are uploaded to online sources, such as Youtube, can have varying frames per seconds, as most of them are collections of short video clips made by the uploader and have different formats. In order to extract motion information from the same time interval on any video, regardless of their frames per second rate, we use seconds as our basic temporal unit. Therefore, our snippets actually correspond to video sequences of lengths in seconds. In the following, we assume that snippets of length seconds are mapped to snippets of length in frames, in order to ease the description of the method.

After calculating the trajectory features for each trajectory $T$, at each position $t = 0 \dots V$, where $V$ is the length of the video, we combine them in snippets. For each snippet, we form *trajectory snippet histograms* to encode the corresponding motion pattern through extracted trajectories.

Consider a snippet $S$ that is centered at frame $s$. We consider all trajectories extracted between $s - \|S\|/2 \leq t \leq s + \|S\|/2$, where $t$ is the ending frame of the trajectory. To

spatially localize the trajectory information, we divide the frames into $N \times N$ spatial grids, and compute histograms for the trajectories whose center points $m_x$ and $m_y$ reside at the corresponding grid. We create 8 bin histograms separately for $l$, $v_x$ and $v_y$ by quantizing corresponding values.

Let's consider $l$, the length of the trajectories, first. Variances in $x$ and $y$ dimensions, $v_x$ and $v_y$, follow a similar process. Let $H_S^l(t)$ be the trajectory snippet histogram for snippet $S$ constructed from the length $l$ of the trajectories that end at frame $t$. It is a vector obtained through concatenating the individual histograms for each spatial grid.

$$H_S^l(t) = (H_S^l(t)_{[1,1]}, \dots H_S^l(t)_{[1,N]}, \dots H_S^l(t)_{[N,N]}) \quad (2)$$

where $H_S^l(t)_{[i,j]}, 0 \leq i, j \leq N,$, is the 8-bin histogram of trajectory lengths, for the trajectories that end at frame $t$ and have $m_x$ and $m_y$ values falling into the $[i,j]^{th}$ grid. For snippet $S$, which is centered at frame $s$, we combine the individual histograms for each $t$, in a single histogram.

$$H_S^l = \sum_{t=s-(\|S\|/2)}^{s+(\|S\|/2)} H_S^l(t) \quad (3)$$

We repeat the same procedure for $v_x$ and $v_y$ to obtain histograms $H_S^x$ and $H_S^y$ respectively. Finally, we combine all of this information for a snippet $S$ as:

$$H_S = (H_S^l, H_S^x, H_S^y) \quad (4)$$

At the end we have a descriptor of $8 \times 3 \times N \times N$ dimensions for each frame $s$ of the video. These descriptors are calculated for each snippet by a sliding window approach.

### 3.3. Classification of usual and unusual videos

We exploit the trajectory snippet histograms for separating unusual videos from usual videos. After extracting the features from each snippet, we use the Bag of Words approach and quantize these histograms into words to generate a *snippet codebook* describing the entire video clip. Then, we train a linear SVM classifier [2] over the training data.

### 3.4. Snapshot discovery

Our goal is then to find the parts of video where the unusual events take place. We call these snippets as *snapshots* corresponding to unusual spatio-temporal video patches.

We address the problem of finding snapshots as finding *discriminative patches* in a video and follow the idea of [8]. However, in our case, a snapshot may include more than a single action unlike [8]. We utilize trajectory snippet histograms to solve this problem.

First, on the training set, for each snippet we find the n-nearest neighbors using trajectory snippet histogram as the feature vector. We check the number of nearest neighbors from usual and unusual videos, and eliminate the snippets with having more neighbors from usual videos. Remaining snippets are used to construct initial models, and an examplar-SVM [12] is trained for each model.

Next, we run our trained models to retrieve similar trajectory snippet histograms for each model. We rank models using two criteria. The first criterion is *appearance consistency*. This is obtained by summing up the top ten SVM detection scores for each model. The second criterion is *purity*. This is calculated by finding the ratio of retrieved features from the unusual videos to the ones from the usual videos. For each model, we linearly combine its *appearance consistency* and *purity* scores. Finally, we rank each model based on the scores, and set the top-ranked models as our unusual video patches.

Alternatively, we also apply an approach very similar to the work in [16] with small differences in implementation. Instead of finding nearest neighbors in the beginning of the algorithm, we cluster the data in the training set into $n/4$ clusters where $n$ is the number of instances. These cluster centers become our initial models, and we test them in the validation set. Models that have less than four firings in the validation set are eliminated, and we train new models using the firings for each model. We test newly trained models in the training set, and follow the same iteration for 5 times. We score each model using their *purity* and *discriminativeness* measures, and retrieve top $T$ models. This method was originally proposed for still images, using HOG features. However, we are easily able to extend into videos by using trajectory snippet histograms as features.

## 4. Experiments

**Datasets:** Videos used in our experiments are downloaded from Youtube, and irrelevant ones are removed manually. We constructed two different datasets. The first set, *Set 1*, has "domain specific" videos. These videos are collected by submitting the query "people falling" for positive videos, and "people dancing", "people walking", "people running" and "people standing" for negative videos. The goal of this set is to test the effectiveness of our method on visually similar usual and unusual videos with low inter-class variations. The second set, *Set 2*, is a more challenging set which consists of videos from variety of activities. Positive videos for this set are retrieved using the query "funny videos", and negative videos are randomly selected. Therefore, there is no restriction on the types of events taking place in videos of *Set 2*. Both sets have 200 positive and 200 negative videos. For each set, we randomly select 60% of videos for the training set, and the remaining 40% for the test set. Both training and test sets are balanced, meaning they have the same amount of positive and negative videos.

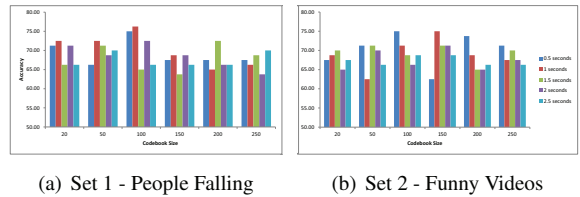

(a) Set 1 - People Falling     (b) Set 2 - Funny Videos

Figure 3. Comparison of performances for trajectory snippet histograms with different snippet lengths and codebook sizes. For both sets, we obtain better results using smaller time snippets.

**Unusual versus usual video classification:** On the task of separating usual videos from unusual videos, we used the snippet codebooks generated from the trajectory snippet histograms. We use BoW approach to quantize descriptors and conduct experiments using different codebook sizes. We also try different snippet lengths. As seen in Figure 3(a), for *Set 1*, using a smaller snippet length gives better results. Note that positive videos in that set consist of people falling, and it makes sense that such action can be seen in snippets of half a second, or one second. Our highest accuracy is 76.25% using a snippet of 1 second and a codebook of size 100. In *Set 2*, since videos can contain any action, we try to learn a more broad definition of unusualness. This is a harder task, but using our descriptor we can still obtain good results, maximum being 75% with snippets of sizes 0.5 and 1 seconds, and codebook size of 100 and 150 words respectively (see Figure 3(b)).

We compare the proposed descriptor based on trajectory snippet histograms with the state-of-the-art descriptors extracted from dense trajectories as used in [19], namely trajectory shape, HOG [3], HOF [11] and MBH [4]. We quantize the features using Bag-of-Words approach. We evaluate codebooks with different sizes, and report the results with highest accuracy values. As shown in Figure 4, the proposed descriptor is competitive with and mostly better than the other descriptors when compared individually. It is not surprising to see that on *Set 1* for "people falling" HOG alone gives the best performance, since the shape information is an important factor for this task. In order to test how much strength is gained with combining different features, we combine all the other descriptors, and also include the snippet histograms as well. The results show that, snippet histograms alone can beat the combination of all other descriptors on *Set 2*, and with the combination of others it becomes the best in both sets. These results show the effectiveness of the proposed descriptor that encodes the motion information in a simple way in capturing the unusualness on many different type of videos.

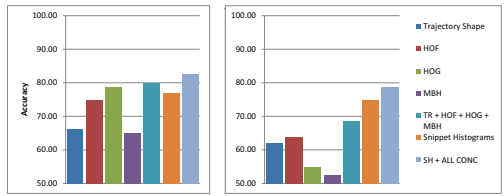

(a) Set 1 - People Falling      (b) Set 2 - Funny Videos

Figure 4. Comparison of our method with state-of-the-art descriptors. As we can observe, the performance of trajectory snippet histograms is better than other descriptors on (b), and it's concatenation with other descriptors gives us the best results in both sets.

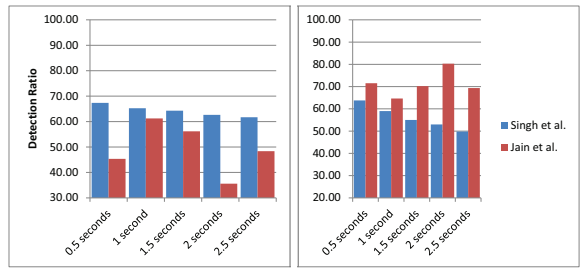

(a) Set 1 - People Falling      (b) Set 2 - Funny Videos

Figure 5. The percentage of firings in positive sets for discriminative snapshots. While using trajectory snippet histograms with [16] gives us better results for *Set 1*, [8] works better in *Set 2*.

As another feature which has been successfully utilized for other problems in the literature, we exploit HOG3D feature [10] on the task of separating usual and unusual videos. However, we could only achieve 73.75% performance on *Set 1* and 65.00% performance on *Set 2* with this feature.

**Discovery of Unusual Video Patches:** With the encouraging results in separation of unusual and usual videos, we then use trajectory snippet histograms to find snapshots as the discriminative video patches in unusual videos. Unlike [8], we do not consider only a subset of spatial grid to find *mid-level discriminative patches*, but consider the trajectory snippet histograms of the entire spatial grid. Over a sliding window approach, with overlapping windows of length *s*, we detect the *discriminative* snippets. Therefore, the output is short snapshots of video where an unusual event occurs.

As seen from some of the snapshots shown in Figure 6, most of the snapshots represent motion patterns with sudden movements. These movements are the results of unexpected events, such as being scared, running into something, being hit by something or falling down. Note that our detector was also able to detect an accidental grenade explosion, which also has sudden movements and long trajectories.

Since the ground truth for snapshots are not available, and difficult to obtain, we use a similar setting as in [5] to quantitatively evaluate the performance of detection. For each snapshot model, the percentage of how many times it was fired in positive videos out of all firings is found. As seen in Figure 5, again the results are better on *Set 2*, compared to *Set 1*.

We compare our descriptor with the HOG3D[10] feature used in [8] using the same setting. We obtain 25.19% on *Set 1* and 30.81% on *Set 2* using the HOG3D feature.

Most of the detected HOG3D snapshots had already been detected by snippet histograms, except for a few like those in the third column of Figure 7. This particular snapshot probably confused snippet histograms as there are people moving around the whole spatial grid. HOG3D descriptors localize features in x and y coordinates, therefore it was able to around the main subject and capture only its motion.

## 5. Conclusion

The problem of detecting unusuality or anomaly has been handled in a very constraint setting up to now. Usually, the video from only one camera is used, so all the actions are seen from one angle only. Most of the works in the literature solve this challenge by detecting irregular events by finding regular events. However, this limits the problem.

Our main goal in this paper is to generalize the solution for the problem described above. We would like to find unusualities in videos, regardless of the scene, actions, or from what angle the video was taken from. This is not an easy task, as we have an infinite number of possible actions, and it would be impossible to learn them all. Furthermore, same action can be seen completely differently in two different perspectives. We propose a simple but efficient method to capture the unusualness in videos, and our experiments give us promising results. As far as we know, this is the first work that attack the problem of discovering unusualness in videos shared in social media regardless of the ongoing events.

## Acknowledgments

This paper is partially supported by TUBITAK grant number 112E174.This paper was also partially supported by the US Department of Defense, the U. S. Army Research Office (W911NF-13-1-0277) and by the National Science Foundation under Grant IIS-0917072. Any opinions, findings, and conclusions or recommendations expressed in this material are those of the author(s) and do not necessarily reflect the views of the National Science Foundation, ARO or the U.S. Government

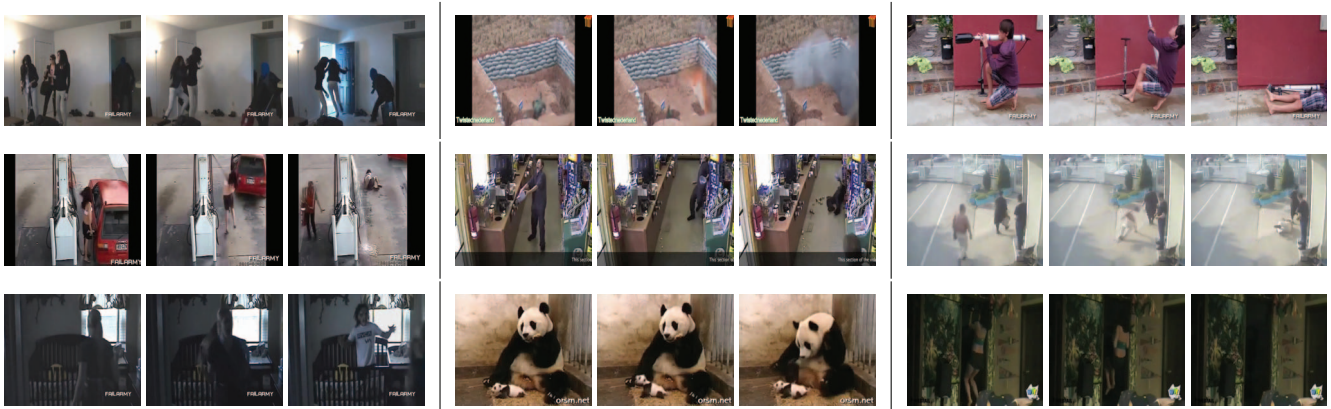

Figure 6. Frames from some of the detected unusual video patches using snippet histograms. As we can see most of the frames contain sudden movements.

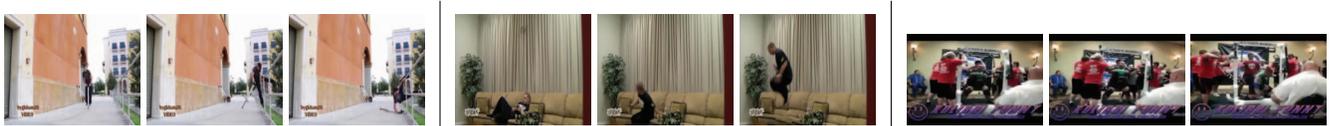

Figure 7. Frames from some of the detected unusual video patches using HOG3D features. Frames on the first two columns were also detected using snippet histograms, while the frames on the third column were only detected by HOG3D features.